\documentclass[times, review, 10pt]{elsarticle}
\usepackage{amssymb}
\usepackage{amsthm}
\usepackage[figuresright]{rotating}
\usepackage{amsmath}
\usepackage{mathrsfs}
\usepackage{newtxmath}
\usepackage{color, colortbl}
\usepackage{amsfonts}
\usepackage{mathtools}
\usepackage{algorithm}
\usepackage{algorithmic}
\usepackage{soul}
\usepackage{url}
\usepackage[utf8]{inputenc}
\usepackage{caption}
\usepackage{subcaption}
\usepackage{graphicx}
\usepackage{booktabs}
\usepackage{xcolor}
\usepackage{multirow}
\usepackage{verbatim}
\usepackage{float}
\usepackage{textcomp}
\usepackage{lipsum}
\usepackage{stfloats}
\usepackage{array}
\usepackage{comment}
\usepackage{makecell}

\usepackage{setspace} 

\usepackage{hyperref}
\hypersetup{
    colorlinks=true,
    linkcolor=blue,
}
\usepackage{caption}
\usepackage{graphicx}
\usepackage{float} 
\usepackage{subcaption}
\journal{Pattern Recognition}
\begin{document}

\begin{frontmatter}

\title{A Renaissance of Explicit Motion Information Mining 
from Transformers for Action Recognition}

\author[inst1,inst2,inst3]{Peiqin Zhuang}
\ead{zhuangpeiqin@pjlab.org.cn}
\author[inst1]{Lei Bai\corref{cor1}}
\cortext[cor1]{Corresponding author}
\ead{bailei@pjlab.org.cn}
\author[inst3]{Yichao Wu}
\author[inst3]{Ding Liang}
\author[inst2]{Luping Zhou}
\author[inst4]{Yali Wang}
\author[inst1]{Wanli Ouyang}

\affiliation[inst1]{organization={Shanghai AI Laboratory},
            city={Shanghai}, country={China}}
\affiliation[inst2]{organization={The University of Sydney},
            city={Sydney}, country={Australia}}
\affiliation[inst3]{organization={SenseTime Group Ltd.},
            city={Beijing}, country={China}}
\affiliation[inst4]{organization={Shenzhen Institute of
Advanced Technology, Chinese Academy of Sciences},
            city={Shenzhen}, country={China}}

\begin{abstract}
Recently, action recognition has been dominated by transformer-based methods, thanks to their spatiotemporal contextual aggregation capacities.
However, despite the significant progress achieved on scene-related datasets, they do not perform well on motion-sensitive datasets due to the lack of elaborate motion modeling designs.
Meanwhile, we observe that the widely-used cost volume in traditional action recognition is highly similar to the affinity matrix defined in self-attention, but equipped with powerful motion modeling capacities. 
In light of this, we propose to integrate those effective motion modeling properties into the existing transformer in a unified and neat way, with the proposal of the Explicit Motion Information Mining module (EMIM). 
In EMIM, we propose to construct the desirable affinity matrix in a cost volume style, where the set of key candidate tokens is sampled from the query-based neighboring area in the next frame in a sliding-window manner. 
Then, the constructed affinity matrix is used to aggregate contextual information for appearance modeling and is converted into motion features for motion modeling as well.
We validate the motion modeling capacities of our method on four widely-used datasets,
and our method performs better than existing state-of-the-art approaches, especially on motion-sensitive datasets, i.e., Something-Something V1 \& V2.
\end{abstract}

\begin{keyword}
Action Recognition, Video Classification, Transformer
\end{keyword}

\end{frontmatter}

\section{Introduction}
\label{introduction}

Action recognition~\cite{tran2015learning, carreira2017quo,lin2019tsm,  tran2018closer,wang2021tdn,zhang2024temporal,ma2024relative,dong2021knowledge} has been a long-term research task in the computer vision community because of its valuable applications
in video surveillance, social media, etc. 
Generally, it is much more challenging than image classification in learning informative representations, as action recognition is required to capture both static spatial context and dynamic temporal relationships
for spatiotemporal representation.

\begin{figure}
     \centering
     \begin{subfigure}[b]{0.35\textwidth}
         \centering
         \includegraphics[width=1.0\textwidth, height=1.28\textwidth]{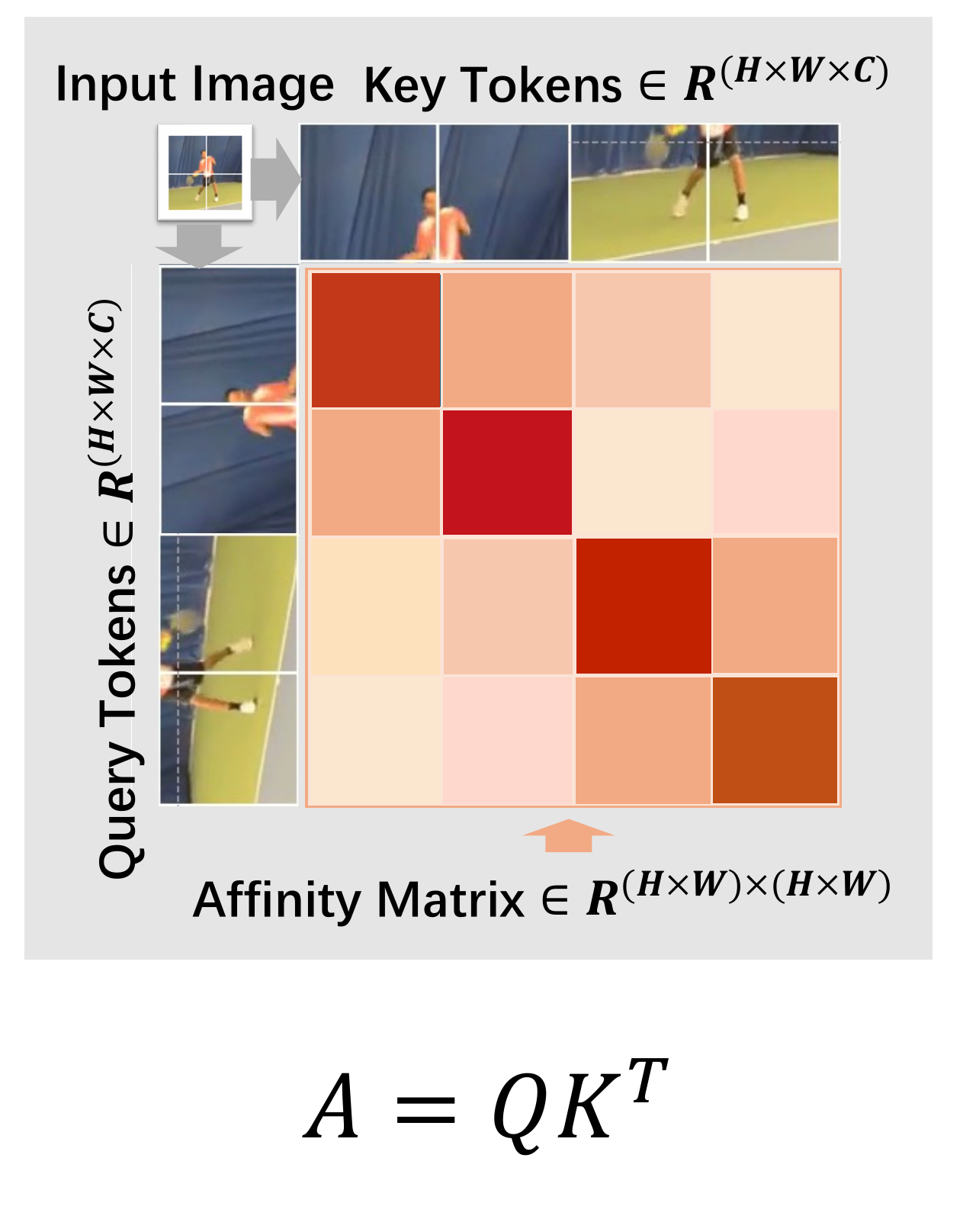}
         \caption{}
         \label{fig:affinity_generation}
     \end{subfigure}
     \hfill
     \begin{subfigure}[b]{0.6\textwidth}
         \centering
         \includegraphics[width=\textwidth, height=0.75\textwidth]{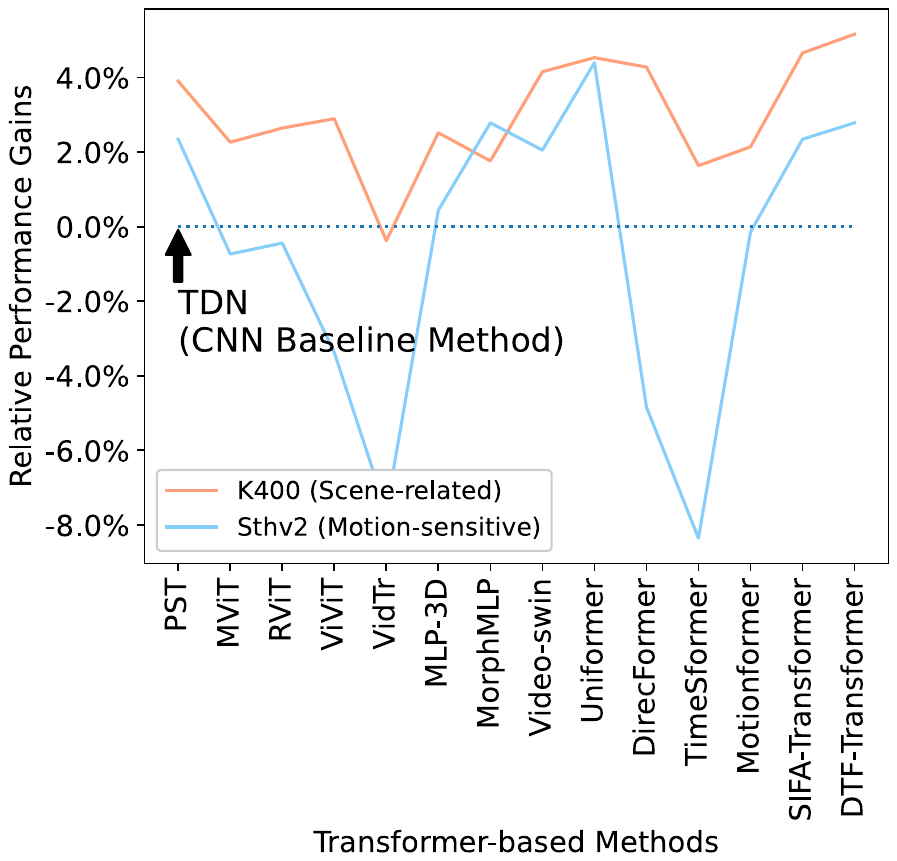}
         \caption{}
         \label{fig:relative_performance_gains}
     \end{subfigure}
        \caption{(a) Affinity matrix generation process. The affinity matrix $\mathbf{A}$ is generated by calculating the visual similarity between the query and key tokens. The depth indicates the magnitude of similarity.  (b) Relative performance gains of the transformer-based methods over the competitive CNN baseline method, i.e., TDN~\cite{wang2021tdn}.
    We choose the competitive TDN as the baseline method. Then, we calculate the relative performance gains of various transformer-based methods on Kinetics-400~\cite{carreira2017quo} and Sthv2~\cite{goyal2017something}, respectively.
    Most of the relative performance gains on Kinetics-400~\cite{carreira2017quo} (scene-related dataset) are smaller than those on Sthv2~\cite{goyal2017something} (motion-sensitive dataset). It means that the current transformer-based methods achieve inferior performance in the motion-sensitive dataset, which requires elaborate designs for effective motion modeling.}
\end{figure}

To this end, many approaches have been proposed by introducing 3D convolutions~\cite{tran2015learning, carreira2017quo}, (2+1)D convolutions~\cite{tran2018closer} and efficient 2D convolutions~\cite{lin2019tsm,wang2021tdn,li2020tea} to capture spatiotemporal information.
Recently, with the success of Vision Transformer (ViT)~\cite{dosovitskiyimage} in image classification,
its variants~\cite{fan2021multiscale,bertasius2021space,patrick2021keeping} have also been developed for action recognition, where self-attention is utilized to aggregate contextual information across spatiotemporal ranges.
Specifically, an affinity matrix $\mathbf{A}$ is first constructed, by performing the inner product between each anchor query token and its potential key candidate tokens,  as shown in Figure~\ref{fig:affinity_generation}.
Then, the calculated affinity matrix $\mathbf{A}$ is used to aggregate similar features from a set of value candidate tokens. 
As a result, the representation power of query tokens would be strengthened with the help of those visually similar tokens,
which consequently boosts the performance.

Despite the remarkable progress achieved by transformer-based methods, we observe that existing video transformers are inferior in modeling complex dynamic temporal relationships.
As shown in Figure~\ref{fig:relative_performance_gains}, the relative performance improvements of video transformers on different datasets vary a lot, where the performance improvements on motion-sensitive datasets\footnote{For motion-sensitive datasets, objects move very quickly across consecutive frames.}, e.g., SSV1 and SSV2~\cite{goyal2017something}\footnote{SSV1 refers to Something-Something V1, while Something-Something V2 is Something-Something V2}, are not as significant as those on scene-related datasets\footnote{For scene-related datasets, action can be recognized by a single image.}, e.g. Kinetics-400~\cite{carreira2017quo}. 
For example, DTF-Transformer~\cite{long2022dynamic} achieves a relative performance gain of $5.1\%$ on Kinetics-400 over the TDN baseline method~\cite{wang2021tdn}, while only obtaining a relative performance gain of $2.8\%$ on SSv2.
Therefore, a question is naturally raised: \textit{Why are transformer-based methods not as effective as expected on motion-sensitive datasets, and how can we address this issue?}
In this paper, we try to answer this question by uncovering potential reasons
and consequently propose our designs to address such issues.

For action recognition, optical flow~\cite{simonyan2014two} has played a key role in depicting the movement status of objects.
Recently, its variant, e.g. cost volume~\cite{kwon2020motionsqueeze,selfy, wang2020video, zhuang2022action} has significantly boosted the performance on motion-sensitive datasets,
thanks to its effective capacity in motion representation.
Generally, cost volume is constructed by measuring the similarities between the query token and its neighboring tokens in the next frame and is directly converted into motion features via feature transformation. 
We observe that the formulation of cost volume is highly similar to the affinity matrix $\mathbf{A}$ defined in self-attention, but differs in several subtle aspects.
We believe such subtle differences may hinder the original self-attention from effective motion modeling.
In this case,
we comprehensively compare self-attention with cost volume 
in Figure~\ref{fig:property_illustration}, and summarize three critical properties of cost volume in effective motion modeling:

\begin{figure*}
    \centering
    \includegraphics[trim=0.3cm 0.35cm 0.1cm 0.2cm, clip, width=0.94\textwidth,height=0.4\textwidth]{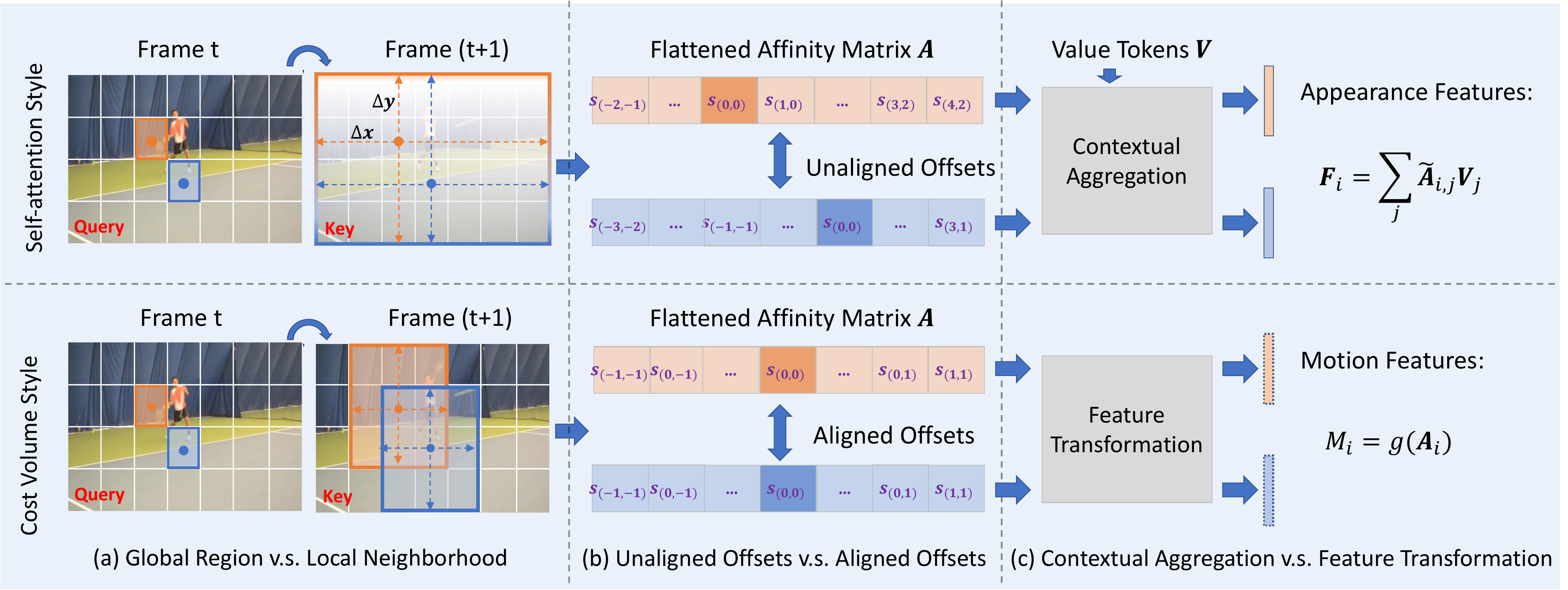}
    \caption{\textbf{Comprehensive comparisons between self-attention and cost volume.} \textbf{(a)} Typically, the neighborhood in the self-attention module is the whole image, while cost volume is calculated within a small local neighborhood.
    \textbf{(b)}
    The set of key tokens is sampled in a \textbf{query-irrelevant} manner in self-attention, while it is sampled from the \textbf{query-based} neighboring area in cost volume.
    \textbf{(c)} The affinity matrix $\mathbf{A}$ in self-attention is used for contextual aggregation with value tokens, while it is converted directly into motion features for motion representation in cost volume.
    $\widetilde{\mathbf{A}}$ denotes a softmax normalized affinity matrix, while $g$ denotes a mapping function for feature transformation. More details can be found in the introduction. (Best viewed when zoomed in and in color)}
    \label{fig:property_illustration}
\end{figure*}

\textbf{1) Local Neighborhood.} For motion-sensitive datasets, 
as objects move very quickly across consecutive frames,
it is important to capture temporal correspondences within a small neighboring region,
rather than aggregating a large amount of contextual information from the global region.
As shown in Figure~\ref{fig:property_illustration} (a), 
for the tennis racket, it is better to focus on its surroundings, rather than the whole image. Much of the background information may even introduce irrelevant or harmful information.
A large part of the background information would not be helpful for accurate action recognition.
In contrast, it may even introduce irrelevant or harmful information in turn.

\textbf{2) Explicit Geometric Inductive Bias.}
Given an anchor query token,
the set of key candidate tokens is sampled from the query-based neighboring area in a sliding-window manner for cost volume generation.
This means that
the offsets of key tokens over their query tokens are definite and well-aligned.
For example, the first and last key tokens would always be located in the upper left and lower right corner of that query-based neighboring area, respectively, no matter what the anchor query token is.
However, this principle is not applicable to traditional self-attention, where every query token shares the same pool of key candidate tokens,
making the offsets of key tokens over their query tokens indefinite, as shown in Figure~\ref{fig:property_illustration} (b).
Definite and well-aligned offsets can indicate how the anchor query token will move toward those pre-defined directions.
In this case, those well-aligned offsets can naturally serve as a kind of explicit motion pattern for later motion modeling, where we call this property the explicit geometric inductive bias.

\textbf{3) Displacement Field Information Encoding.} 
In the traditional self-attention module, the calculated affinity matrix is generally used for contextual aggregation to enrich the representation power of features.
In this case, the displacement field information of key tokens over their anchor query token,
i.e., offsets and their corresponding similarities/probabilities,
would ultimately be lost.
Although positional embeddings and relative position bias have been proposed to alleviate this situation,
such implicit offset information injections may still be far from satisfactory in motion-sensitive scenarios.
Alternatively,
the constructed cost volume is directly converted into motion features,
where the displacement field information would be encoded into motion features for discriminative motion representation. Such motion features are complementary to the appearance features obtained by contextual aggregation.

After keeping those essential properties in mind, we propose to mine explicit motion information from the existing transformer for better motion capturing.
We achieve this goal with the proposal of the Explicit Motion Information Mining module (EMIM).
The EMIM module is based on the traditional self-attention module but is elaborately designed in terms of the aforementioned properties.
In EMIM, we propose to construct the desirable affinity matrix in a cost-volume style.
Concretely, for each query token, the set of key candidate tokens will be sampled from the query-based neighboring area in the next frame in a sliding-window manner, rather than sampled in a query-irrelevant manner.
Then, 
the calculated affinity matrix is used for contextual aggregation with value candidate tokens
and converted into a motion representation via feature transformation.
As a result,
the proposed EMIM module can preserve the original contextual aggregation capacity in self-attention for appearance modeling and develop a new capacity for motion modeling.
We evaluate the effectiveness of our method on four widely-used datasets.
As expected,
our method performs better than existing state-of-the-art approaches, especially on motion-sensitive datasets,
e.g. SSV1 \& SSV2, which demotes the motion modeling capacities of our method.

In summary, our contributions are summarized as follows:
\begin{itemize}
    \item We comprehensively compare the differences between cost volume and self-attention. Furthermore, 
    we summarize and empirically validate three key properties of cost volume in effective motion modeling. 
    \item We propose the Explicit Motion Information Mining module (EMIM) to enrich the motion modeling capacity for the existing transformer. The EMIM module preserves the original capacity for contextual aggregation and develops a new ability in effective motion modeling.
    \item We validate the effectiveness of the proposed method on several widely-used datasets, and achieve state-of-the-art performances, especially on motion-sensitive datasets,
    demonstrating the effectiveness of our designs.
\end{itemize}

\section{Related Work}

\subsection{CNN-based Approaches}
With the emergence of deep neural networks, CNN-based approaches~\cite{tran2015learning, carreira2017quo,lin2019tsm,  tran2018closer, wang2021tdn} have gradually become the norm in video classification, compared to hand-crafted feature-based methods~\cite{wang2013dense}.
Generally,
these CNN-based approaches could be roughly categorized into 3D CNNs and 2D CNNs.

3D CNNs and their variants have been proposed to simultaneously capture spatial and temporal information. Among them, C3D~\cite{tran2015learning} first proposes utilizing 3D convolution to aggregate spatiotemporal information within cubes, to strengthen the spatiotemporal modeling capacities of CNNs. 
After that, many approaches have been proposed to alleviate the initialization problem~\cite{carreira2017quo} or perform spatiotemporal decomposition~\cite{qiu2017learning,tran2018closer}.

Despite promising performances achieved by 3D CNNs,
they are still computationally intensive compared to 2D CNNs. 
For 2D CNNs,
frame-wise features are first extracted by computationally efficient 2D convolution. Then, various temporal designs are built on top of frame-wise features to further extract distinct motion representations,  either short-range or long-term.
For example, TSM~\cite{lin2019tsm} proposes integrating motion information from adjacent frames by performing channel-wise shifts along the temporal dimension.
Besides that, several approaches~\cite{wang2021tdn, li2020tea} have been proposed by exploiting the feature differences between adjacent frames and distant frames to build complicated and dynamic temporal relationships between frames in a video.
In addition, MLENet~\cite{wang2024mlenet} proposes to utilize optical flow-guided features extracted from a 2D CNN to enhance local details.

As it is shown, capturing dynamic temporal relationships among frames always plays a key role in the design of 3D CNNs and 2D CNNs. Our work also follows the important spirit of motion modeling in CNNs. 
However,  we focus on integrating motion modeling into transformer-based models, which has been less explored in transformer-based models.

\subsection{Transformer-based Approaches}
Over the past three years, the transformer has become dominant in image classification~\cite{dosovitskiyimage}, object detection~\cite{carion2020end}, semantic segmentation~\cite{strudel2021segmenter} and etc. Meanwhile, it has also made remarkable progress in action recognition~\cite{bertasius2021space, liu2022video, carreira2017quo, goyal2017something, ma2024relative}, thanks to its ability to capture global dependencies across spatial and temporal ranges for contextual aggregation.
However, the native application of self-attention in the video domain brings heavy computational costs, due to the large number of spatiotemporal tokens.

To this end, many approaches have been proposed to reduce computational costs by decomposing the original spatiotemporal self-attention into a spatial self-attention and a temporal self-attention~\cite{bertasius2021space, arnab2021vivit}, reducing the length of key and value tokens by performing pooling operations~\cite{fan2021multiscale}, performing self-attention within 3D local windows~\cite{liu2022video}, introducing relative-position embedding based patch tokenization~\cite{ma2024relative} and etc.

Previous transformer-based methods mainly focus on how to aggregate contextual information across spatial and temporal ranges with minimum computational costs.
However, such a contextual aggregation paradigm may be only beneficial for scene-related datasets, like Kinetics-400~\cite{carreira2017quo}, but not for motion-sensitive datasets, e.g., SSV1 \& SSV2~\cite{goyal2017something}, where motion-sensitive datasets require explicit and more elaborate designs to capture dynamic movements in a video. 
Such an important factor has been ignored in previous transformer-based methods.
This factor has been ignored in previous transformer-based methods.
To this end, we propose to enrich the motion modeling capacity in naive transformers. We make an attempt to preserve the original contextual aggregation capacity in appearance modeling while developing a new ability in effective motion modeling.

\subsection{Foundation Model-based Approaches}
There is a trend of foundation model–based approaches, which propose to train a vision encoder on large-scale data, and subsequently use the pretrained encoder for fine-tuning on a variety of downstream datasets. One line of research~\cite{tong2022videomae, feichtenhofer2022masked, wang2023videomae,huang2023mgmae, li2023unmasked} resorts to masking strategies to implicitly learn complex spatiotemporal relationships from video cubes. For example, \cite{tong2022videomae} and ~\cite{feichtenhofer2022masked} propose to randomly mask a proportion of video tokens to improve the model’s ability to predict the missing content.
Furthermore, VideoMAE V2~\cite{wang2023videomae} proposes a dual masking strategy to accelerate the pre-training procedure, particularly when models are required to be scaled up. 
Meanwhile,
another line of research advocates employing the multimodal CLIP model~\cite{radford2021learning} pre-trained on large-scale image–text data, and adapting it to the video domain through prompt tuning~\cite{wang2025vlpa}, adapter modules~\cite{wang2024multimodal, lin2022frozen, yang2023aim}, or LoRA strategies~\cite{li2024zeroi2v} or textual guidance~\cite{wu2023cap4video, wu2024cap4video++}.
To sum up, these methods focus on how to adopt various masking strategies to learn spatiotemporal relationships in an implicit manner or bridge the domain gap with minimal additional parameters.
However, they do not introduce explicit temporal modeling designs into the vanilla ViT. Instead, we propose to perform explicit motion modeling to enhance the temporal modeling capacity of the transformer-based architectures.

\subsection{Cost Volume}
Cost volume is obtained by calculating visual similarities from pairs of pixels.
Given an anchor pixel, a set of candidate pixels will be sampled from the anchor-based neighborhood in the next frame.
The visual similarities could be further utilized to depict how an object in the current frame would move in the next frame.
Based on that, it has been exploited for displacement estimation in optical flow estimation~\cite{dosovitskiy2015flownet}, stereo matching~\cite{zbontar2016stereo}, and depth estimation~\cite{imdpsnet}.
Recently, it has also been introduced to action recognition with CNN-based methods~\cite{kwon2020motionsqueeze,selfy,wang2020video,zhuang2022action}, where the calculated cost volume is used to improve motion modeling capacity, accompanied by appearance modeling in CNN-based methods. 
For example,
MSNet~\cite{kwon2020motionsqueeze} proposes to use the calculated cost volume to estimate the displacement field along the horizontal and vertical dimensions, while 
SELFY~\cite{selfy} exploits various feature extraction methods to extract generalized motion representations from the aforementioned cost volume for robust action recognition.

Although cost volume shares the merit in effective motion modeling
with CNN-based methods,
it has never been exploited with transformer-based methods. Meanwhile, we also notice that the cost volume is highly similar to the affinity matrix defined in self-attention. Based on that, we propose to mine those effective motion modeling properties from cost volume and integrate those properties into the existing transformer, to enrich the motion modeling capacity.

\textbf{Discussion.} We notice that there is concurrent work in video frame interpolation, i.e., EMA~\cite{zhang2023extracting}, which is similar to our work. 
Although EMA and our work share the same spirit of reuse of the affinity matrix in the transformer, our work differs from EMA in three aspects: 1) Different Fields. 2) Token Sampling Strategy. EMA adopts the non-sliding window strategy for key/value tokens sampling, while our work proposes to sample key/value tokens from the query-based neighboring area in a sliding-window manner.
Our sampling strategy shares the merit of having aligned offsets for each anchor query token, which could naturally serve a kind of explicit motion pattern for later motion modeling. We will investigate the influence of different token sampling strategies in our ablation studies in Section~\ref{section_ablation_study}. 3) Different uses of the affinity matrix. EMA utilizes the affinity matrix to perform a weighted sum over a coordinate map to estimate displacements, which is very similar to contextual aggregation in self-attention. However, we propose to directly convert the affinity matrix into motion features, in order to encode the displacement field information for discriminative motion representation.

\section{Method}
\subsection{Overview}
\begin{figure*}[]
    \centering
    \includegraphics[trim=0.1cm 0.2cm 0.3cm 0.3cm, clip, width=0.94\textwidth,height=0.50\textwidth]{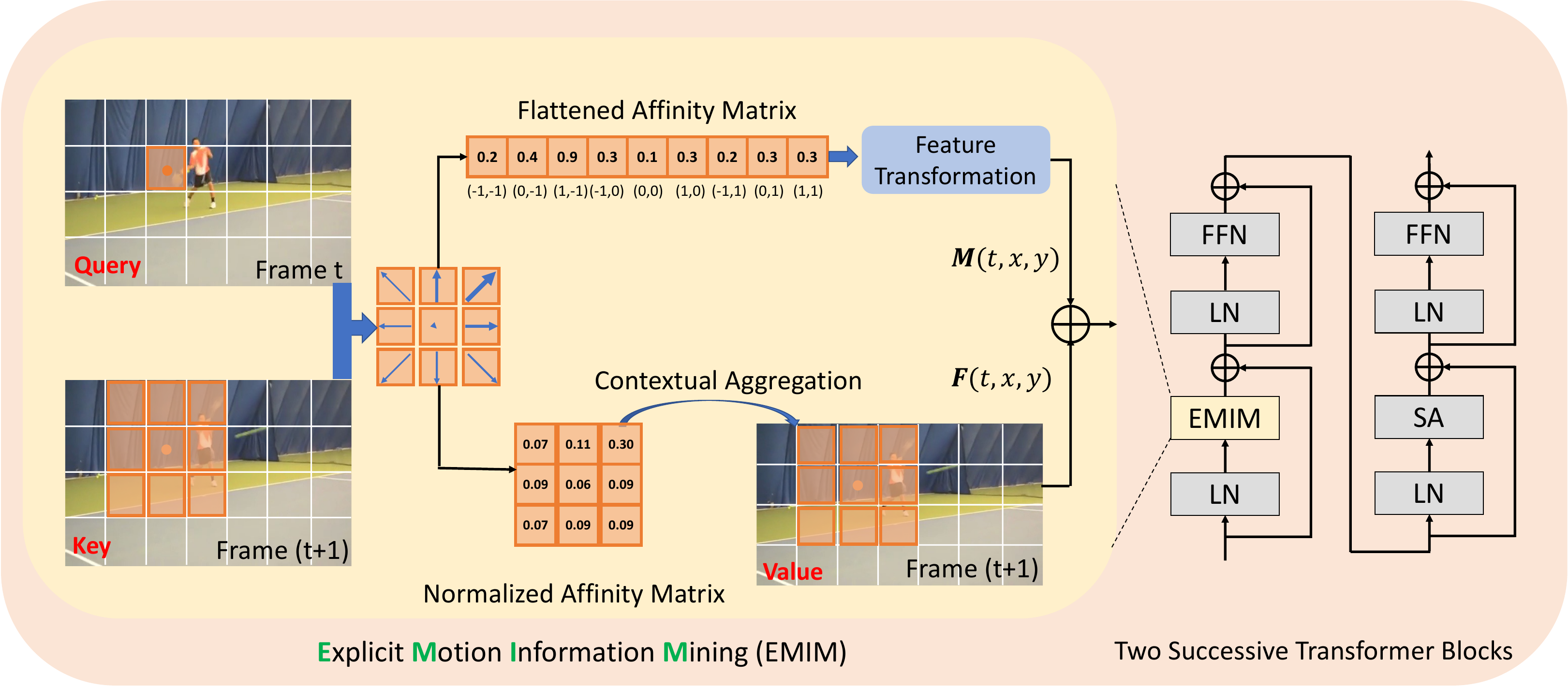}
    \caption{\textbf{Explicit Motion Information Mining module (EMIM).}
    We replace the self-attention module with our EMIM in the first transformer block for every two successive transformer blocks.
    Given an anchor query token $\mathbf{Q}(t,x,y)$ at the position of $(x,y)$ in the $t^{th}$ frame, a set of key candidate tokens would be adaptively sampled from the \textbf{query-based} neighboring area in the $(t+1)^{th}$ frame to construct the affinity matrix.
    After that, on the one hand, the constructed affinity matrix is utilized for contextual aggregation for appearance modeling, i.e., $\mathbf{F}(t,x,y)$; on the other hand, it will be flattened and converted into motion features via feature transformation for motion modeling, i.e. $\mathbf{M}(t,x,y)$. Finally, these two kinds of complementary features are combined and fed into the following layers. (Better viewed when zoomed in and in color)
    }
    \label{fig:framework}
\end{figure*}

As shown in Figure~\ref{fig:framework}, 
for every two successive transformer blocks, we replace the self-attention module in the first transformer block with our proposed explicit motion information mining module (EMIM) and keep the second transformer block unchanged.
The proposed EMIM module conducts the following operations.

1) Given an anchor query token $\mathbf{Q}(t,x,y)$ at the position of $(x,y)$ in the $t^{th}$ frame, a set of key candidate tokens would be sampled from the query-based area in the $(t+1)^{th}$ frame for the affinity matrix construction.

2)  Then, the constructed affinity matrix is both utilized for contextual aggregation for appearance modeling, i.e., $\mathbf{F}(t,x,y)$, and converted into motion features via feature transformation for motion modeling, i.e., $\mathbf{M}(t,x,y)$.

3) Finally, these two types of complementary features are combined and fed into the following layers.

\subsection{Transformer Block}
Given a video clip, it is divided into a set of non-overlapping patches.
Then those patches are flattened and converted into a set of tokens $\mathbf{X} \in \mathbb{R}^{N \times C}$, where
 N is the number of tokens and $C$ denotes the dimension of each token.
 After that, those tokens are fed into transformer blocks layer by layer.

 Suppose that the transformer backbone has $L$ layers of transformer blocks. Generally, a transformer block consists of a Self-attention (SA) module, Layer Normalization (LN), and Fast-Forward Networks (FFN).
 Given an input tensor $\mathbf{Z}^{l-1}$ for the transformer block in layer $l$, it is calculated with the following steps:
  \begin{equation}
     \begin{aligned}
   &  \mathbf{Y}^{l} = \textbf{SA}(\textbf{LN}(\mathbf{Z}^{l-1})) + \mathbf{Z}^{l-1}, \\
   &  \mathbf{Z}^{l} = \textbf{FFN}(\textbf{LN}(\mathbf{Y}^{l})) + \mathbf{Y}^{l},
     \end{aligned}
 \end{equation}
 where $\mathbf{Y}^{l}$ and $\mathbf{Z}^{l}$ denote the output tensor obtained after the SA and FFN modules, respectively.
 In the SA module, the input tensor $\mathbf{Z}^{l-1}$ is converted into $\mathbf{Q}$, $\mathbf{K}$, $\mathbf{V} \in \mathbb{R}^{N \times d^{l}}$ to represent the \textit{query}, \textit{key} and \textit{value} tokens, respectively. $d^{l}$ is the dimension of the tokens on layer $l$.
 Then, those tokens are operated in the following way for contextual aggregation  (We omit multi-head and LN here for simplicity):
 \begin{equation}
     \textbf{SA}(\mathbf{Z}^{l-1}) = \textbf{SoftMax}(\mathbf{Q} \mathbf{K}^{\top} / \sqrt{d^{l}})\mathbf{V},
 \end{equation}
where $\mathbf{A} = \frac{\mathbf{Q}\mathbf{K}^{\top}}{\sqrt{d^{l}}} \in \mathbb{R}^{N \times N}$ denotes the affinity matrix obtained by performing the scaled inner product between the query tokens and the key tokens.
The scaling factor of $\sqrt{d^l}$ is introduced to prevent the vanishing gradient problem due to large values.
Each row in the affinity matrix indicates the visual similarities between an anchor query token and a pool of key candidate tokens.
Therefore, it could be used to strengthen the representation power of a given anchor key token with its visually similar features, in terms of the magnitude of visual similarities.
Note that the sampling strategy of key candidate tokens is query-invariant, as all query tokens share the same pool of candidate tokens in the traditional transformer blocks.
Since tokens in a video clip span different spatial and temporal ranges, such a contextual aggregation would make it easy for the model to correctly recognize actions by capturing long-term dependencies across the whole video clip.

~\subsection{Explicit Motion Information Mining}
Although the traditional self-attention module is capable of capturing appearance information through contextual aggregation, it may suffer from effectively capturing motion information due to the lack of elaborate motion modeling designs.
In this case, we propose to take a further step to mine explicit motion information from the traditional self-attention module,
where we name our proposed module \textbf{E}xplicit \textbf{M}otion \textbf{I}nformation \textbf{M}ining module (EMIM).

\textbf{Affinity Matrix Construction}. The key to the EMIM module lies in how to construct the desirable affinity matrix for motion modeling.
In practice, we propose to construct the affinity matrix as follows:
\begin{equation}
\small
\begin{split}
    \mathbf{A}&= \{\frac{\mathbf{Q}(t, x, y)\mathbf{K}^{\top}(t+1, x + \Delta x, y + \Delta y)}{\sqrt{d^{l}}} + \mathbf{B}(\Delta x, \Delta y) \;| \\
    & \hspace{2em} t \in \{1,2, \dots, T\}, \\
    & \hspace{2em} x \in \{1,2, \dots, S\}, \;\;\;\; y \in \{1,2, \dots, S\}, \\
    & \hspace{2em} \Delta x \in \{-P, \dots, P\}, \Delta y \in \{-P, \dots, P\}
    \}.
\end{split}
\end{equation}

Given the query, key, and value tokens, we first reshape those tokens into token volumes, i.e. $\mathbf{Q}, \mathbf{K}, \mathbf{V} \in \mathbb{R}^{(T \times S \times S) \times C}$, 
where S and T denote the spatial size and temporal size of the token volumes, respectively.
$\mathbf{Q}(t, x, y)$ denotes a token sampled from the spatial position $(x, y)$ in the $t^{th}$ frame  within the volume of the token $\mathbf{Q}$.
For each anchor query token $\mathbf{Q}(t, x, y)$, a set of key candidate tokens $\mathbf{K}(t+1, x + \Delta x, y + \Delta y)$ would be adaptively sampled from the query-based neighboring area $(x + \Delta x, y + \Delta y)$ in the $(t+1)^{th}$ frame. 
$\Delta x$ and $\Delta y$ denote the displacements of the key candidate tokens over the anchor query token, and range from $[-P, P]$, where P is the maximum displacement. 
Typically, the maximum displacement $P$ is smaller than the spatial size S, to precisely capture local movements.
We empirically set it as $7$ by default.
We pad the boundary with a constant value, i.e., 1e-6, to extend the boundary, in the case of $x + \Delta x$ or $y + \Delta y$ extending the boundary. 
In addition to the scaled inner production,
we follow Swin transformer's~\cite{liu2021swin} strategy to add a relative positional bias matrix $\mathbf{B} \in \mathbb{R}^{(2P+1) \times (2P+1)}$ for better affinity estimation.
The relative positional bias can be seen as an implicit geometric inductive bias, further enhancing the effect of geometric inductive bias for neighborhood structure modeling.
The resulting affinity matrix $\mathbf{A} \in \mathbb{R}^{(T \times S \times S) \times (2P+1)^2}$ represents the visual similarities between anchor query tokens and their corresponding key tokens sampled from the query-based neighborhood in the next frame.

\textbf{Contextual Aggregation.}
After obtaining the affinity matrix $\mathbf{A}$, we perform the softmax operation to obtain the normalized affinity matrix $\widetilde{\mathbf{A}}$.
Then, we use $\widetilde{\mathbf{A}}$ to perform contextual aggregation with the corresponding value tokens for the appearance modeling. 
The resulting appearance feature $\mathbf{F} \in \mathbb{R}^{(T \times S \times S) \times d^{l}}$ is computed as follows:
\begin{equation}
\small
\begin{split}
    \mathbf{F}(t, x, y) &= \sum_{\Delta x, \Delta y} \widetilde{\mathbf{A}}(t,x,y, \Delta x, \Delta y) \mathbf{V}^{\top}(t+1, x + \Delta x, y + \Delta y).
\end{split}
\end{equation}
For each anchor query token $\mathbf{Q}(t,x,y)$ in the $t^{th}$ frame,
the set of value candidate tokens $\mathbf{V}(t+1, x + \Delta x, y + \Delta y)$  would be sampled from the query-based neighboring area in the $(t+1)^{th}$ frame.
The resulting appearance feature $\mathbf{F}(t,x,y)$ is obtained by averaging every value token with its corresponding affinity value, which is a form of the weighted sum of value tokens.

\textbf{Motion Modeling.}
With the affinity matrix obtained, we also utilize it for motion representation. Since each row in the affinity matrix represents how the anchor query token would move to the set of key candidate tokens toward pre-defined directions in the next frame, 
it naturally serves as a kind of explicit motion pattern in motion modeling. 
The resulting motion feature $\mathbf{M} \in \mathbb{R}^{(T \times S \times S) \times d^{l}}$ is obtained as follows:
\begin{equation}
    \mathbf{M}(t,x,y) = f(\mathbf{A}(t,x,y)),
\end{equation}
where $f$ is a mapping function for feature transformation. It is composed of FC($(2P+1)^2$, $(2P+1)^2 \times 4$), GELU and  FC($(2P+1)^2 \times 4$, $d^l$).
With the mapping function $f$,
we can encode these explicit motion patterns into informative motion representations, which is complementary to the appearance features.

After that, we combine both the appearance feature $\mathbf{F}$ and the motion feature $\mathbf{M}$ and feed the combined feature into the following layers.

\section{Experiments}
\subsection{Datasets}
To verify the effectiveness of our proposed method, we conducted extensive experiments on four widely used datasets, including Kinetics-400~\cite{carreira2017quo}, SSV1 \& SSV2~\cite{goyal2017something}, and Diving-48 V2~\cite{li2018resound}. 
To be specific,
Kinetics-400 is a 400-classes dataset with 240k training videos and 20k validation videos. Most of its videos are scene-based and could be easily recognized without elaborate motion modeling designs.
SSV1 \& SSV2 are 174-classes motion-sensitive datasets, containing $108,499$ and $220,847$ videos, respectively. 
Diving-48 V2 is a collection of diving activities with $15,027$ training videos and $1,970$ validation videos. It is required to capture long-term temporal dynamics throughout the whole video.
Note that since SSV1 \& SSV2 and Diving-48 V2 are motion-sensitive datasets, they could be used to examine the motion modeling capacities of our proposed method.
In this case, we mainly focus on performance improvements on these datasets.

\subsection{Implementaion Details}
\textbf{Backbone Selection.} We choose Uniformer~\cite{liuniformer} as our backbone because of its competitive performance.
Uniformer is a hybrid architecture, consisting of convolution blocks in the shadow layers (Stage 1, 2) and transformer blocks in the deep layers (Stage 3, 4).
For every two successive transformer blocks in Stage 3 and 4, we replace the original self-attention module in the first transformer block with our proposed EMIM module and keep the second transformer block unchanged.
We empirically find that this kind of replacement strategy performs best. 
In addition, 
we also remove the temporal downsampling strategy in Stage 1 to preserve more temporal information.

\textbf{Training}. Unless stated otherwise, we train our model with the following settings.
For frame sampling, 
we sample frames from Kinetics-400 videos using a dense sampling strategy,
while the sparse sampling strategy is adopted for other datasets.
The short side of the frames is randomly scaled between $256$ and $320$, and the patches are cropped from those scaled frames with the size of $224 \times 224$. We optimize our models with AdamW optimizer, the momentum of $0.9$, and the weight decay of $0.05$.The initial learning rate for Kinetics-400 and other datasets is $0.0001$ and $0.0002$, respectively, and the batch size is $32$.
The training process for Kinetics-400 lasts for 110 epochs with 10 warm-up epochs. For SSV1, the training epochs are 50 with 5 warm-up epochs. For Some-Something V2 and Diving-48 V2, the training epochs are 60 with 5 warm-up epochs.
To speed up the sliding window sampling strategy in implementation, we adopt the NATTEN package~\cite{hassani2022neighborhood}.

\textbf{Testing.} We follow the widely-used inference protocol in most transformer-based models. 
The short side of the frames is fixed at $224$ for later cropping. In general, 3 patches (left, top, right) would be cropped for each frame. 
For Kinetics-400, we uniformly sample 4 video clips from the whole video with a dense sampling strategy, and average their prediction results for video prediction. 
For other datasets, we uniformly sample $16$ or $32$ frames from the whole video to form a video clip for video prediction.
\subsection{Comparison with State-of-The-Art Methods}

\begin{table*}[]
    \centering
    \resizebox{1.0\textwidth}{0.36\textwidth}{
    \begin{tabular}{c|c|c|c|c|c|c|c}
    \toprule[1pt] 
     Method  &  Pretrained &  Frames & GFlops $\times$ views & V1-Top1 & V1-Top5 & V2-Top1 & V2-Top5 \\
    \hline
    TSM R50~\cite{lin2019tsm} & ImageNet-1K & $16$ & $65 \times 1 \times 1$ & $47.2$ & $77.1$ & $63.4$ & $88.5$ \\
     TEA R50~\cite{li2020tea} & ImageNet-1K & $16$ & $70 \times 10 \times 3$ & $52.3$ & $81.9$ & N/A & N/A \\
    MSNet R50~\cite{kwon2020motionsqueeze} & ImageNet-1K & $8+16$  & $101 \times 1 \times 1$ & $55.1$ & $84.0$ & $67.1$ & $91.0$ \\
    MoDS R50~\cite{zhuang2022action} & ImageNet-1K & $8+16$ & $106 \times 1 \times 1$ & $56.6$ & $84.0$ & $67.1$ & $90.7$ \\
    SELFYNet R50~\cite{kwon2020motionsqueeze} & ImageNet-1K & $8+16$ & $114 \times 2 \times 1$ & $56.6$ & $84.4$ & $67.7$ & $91.1$ \\
    TDN R101~\cite{wang2021tdn} & ImageNet-1K & $8+16$ & $198 \times 1 \times 3$ & $56.8$ & $82.9$ & $68.2$ & $91.6$ \\
   \hline
      TimeSformer-HR~\cite{bertasius2021space} & ImageNet-21K & $16$ & $1703 \times 1 \times 3$ & N/A & N/A & $62.5$ & N/A \\ 
         Motionformer-L~\cite{patrick2021keeping} & Kinetics-400 & $32$ & $1185.1 \times 1 \times 3$ & N/A & N/A & $68.1$ & $91.2$ \\
         Video Swin~\cite{liu2022video} & Kinetics-400 & $32$ & $321 \times 1 \times 3$ & N/A & N/A & $69.6$ & $92.7$ \\
         PST-B ~\cite{xiang2022spatiotemporal} & Kinetics-400 & $32$ & $252 \times 1 \times 3$ & $58.3$ & $83.9$ & $69.8$ & $93.0$ \\
       SIFA-Transformer~\cite{long2022stand} & ImageNet-21K & $32$ & $270 \times 1 \times 3$ & $57.3$ & $85.1$ & $69.8$ & $93.1$ \\
         DTF-Transformer~\cite{long2022dynamic} & ImageNet-21K & $32$ & $266 \times 1 \times 3$ & $57.9$ & $85.7$ & $70.1$ & $93.2$ \\
         MorphMLP-B~\cite{zhang2021morphmlp} & ImageNet-1K & $32$ & $197 \times 1 \times 3$ & $57.4$ & $84.5$ & $70.1$ & $92.8$ \\
    MViT-B~\cite{fan2021multiscale} & Kinetics-400 & $64$  & $455 \times 1 \times 3$ & N/A & N/A & $67.7$ & $90.9$ \\
         MLP-3D-L~\cite{qiu2022mlp} & ImageNet-1K & $64$ & $336 \times 1 \times 3$ & $56.5$ & $83.5$  & $68.5$ & $92.0$ \\
     Unifomer-B~\cite{liuniformer} & Kinetics-400 & $32$ & $259 \times 1 \times 3$ & $61.0$ & $87.6$ & $71.2$ & 
    $92.8$ \\ 
     \hline 
  EMIM-S & Kinetics-400 & $16$ & $85 \times 1 \times 3$ & $59.9$ & $86.5$ & $71.3$ & $93.6$ \\
  EMIM-B & Kinetics-400 &  $16$ & $198 \times 1 \times 3$ & $60.8$ & $87.8$ & $72.5$ & $93.7$ \\
    EMIM-B & Kinetics-400 & $32$ &  $526 \times 1 \times 3$ & $\mathbf{62.4}$  & $\mathbf{88.0}$ &  $\textbf{73.2}$ & $\mathbf{93.9}$ \\ 
    \bottomrule[1pt] 
    \end{tabular}
    }
    \caption{Comparison with state-of-the-art methods on SSV1 \& SSV2}
    \label{ssv_result}
\end{table*}

\textbf{Evaluation on Something-Something V1 \& V2\footnote{EMIM-S and EMIM-B differ in the depths of blocks, i.e., \{3,4,8,3\} v.s. \{5,8,20,7\} } }.
    We employ our proposed method on both Something-Something V1 \& V2, to verify the capacity of motion modeling. The comparison results between our method and other state-of-the-art methods are shown in Table~\ref{ssv_result}.
    For a clear illustration,
    we group those state-of-the-art methods in terms of different backbone types, pre-trained datasets, frame numbers and etc. 
    and achieves $62.4\%$ on V1 and $73.2\%$ on V2, respectively.
    Our method performs better than those spatiotemporal self-attention decomposition approaches~\cite{bertasius2021space}. It indicates that, in addition to taking into account the computational complexity in the design of spatiotemporal self-attention, 
    it is also crucial to pay more attention to the motion modeling capacity in self-attention.
    Also, our method performs better than those approaches~\cite {long2022dynamic,long2022stand}, e.g., SIFA-Transformer, where key/value tokens are also sampled from the next frame in a sliding-window manner.
    However, the affinity matrix constructed in these approaches is used for contextual aggregation only. 
    In this case, the displacement field information of key candidate tokens over the anchor query token,
    i.e., offsets and their corresponding probabilities,
    would be lost in the end. Therefore, it proves that it is important to preserve the displacement field information obtained within the affinity matrix and to make full use of the valuable information for discriminative motion representations.

\begin{table}[]
    \centering
    \resizebox{0.6\textwidth}{0.13\textwidth}{
    \begin{tabular}{c|c|c}
    \hline \hline
        Methods & Pretrained  & Top1 \\ \hline
       TDN~\cite{wang2021tdn} & ImageNet-1K &   80.5 \\
      TimeSformer-L~\cite{bertasius2021space}  & ImageNet-21K  & $81.0$ \\
       PST-B~\cite{xiang2022spatiotemporal} & Kinetics-400 & 86.0 \\
        ORVIT~\cite{herzig2022object} & ImageNet-21K & $88.0$ \\
       \hline
       EMIM (Our Method) & ImageNet-1K & $\textbf{91.6}$ \\
       EMIM (Our Method) & Kinetics-400 & $\textbf{92.2}$ \\ 
       \hline \hline
     \end{tabular}
     }
    \caption{Comparison with state-of-the-art methods on Diving-48}
    \label{diving}
\end{table}

\textbf{Evaluation on Diving-48.}
We also compare our method with other state-of-the-art methods on Diving-48. There exists a large number of motion dynamics throughout the whole video in Diving-48, which is helpful for examining the capacity of motion modeling. As shown in Table~\ref{diving}, our method surpasses state-of-the-art methods on Diving-48. For example, our method performs better than the recent advanced transformer-based method, i.e., ORVIT~\cite{herzig2022object}, which uses auxiliary trajectory information to capture movements.
ORVIT is also a motion-dedicated transformer method, which uses auxiliary trajectory information obtained by a tracking model, to capture motion information.
In this case, the competitive result of our method verifies the strong motion modeling capacity with our elaborate designs.

\begin{table*}[]
    \centering
    \resizebox{1.0\textwidth}{0.30\textwidth}{
    \begin{tabular}{c|c|c|c|c|c}
    \toprule[1pt] 
      Method  &  Pretrained & Frames &  GFlops $\times$ views & Top1 & Top5 \\
    \hline
    TSM R50~\cite{lin2019tsm} & ImageNet-1K & $8$ & $33 \times 10 \times 1$ & $74.1$ & $91.2$ \\
    MoDS R50~\cite{zhuang2022action} & ImageNet-1K & $8$ & $37 \times 10 \times 3$ & $75.7$ & N/A \\
    TEA R50~\cite{li2020tea} & ImageNet-1K & $16$ & $70 \times 10 \times 3$ & $76.1$ & $92.5$ \\
     MSNet R50~\cite{kwon2020motionsqueeze} & ImageNet-1K & $16$ & $67 \times 10 \times 1$ & $76.4$ & $77.4$ \\
     SELFYNet R50~\cite{selfy} & ImageNet-1K & $16$ & $77 \times 10 \times 3$ & $77.1$ & N/A \\
     SlowFast R101~\cite{feichtenhofer2019slowfast} & N/A & $16$ & $234 \times 10 \times 3$ & $79.8$ & $93.9$ \\
   TDN R101~\cite{wang2021tdn} & ImageNet-1K & $8+16$ & $198 \times 10 \times 3$ & $79.4$ & $94.4$ \\
   \hline
     Motionformer-HR~\cite{patrick2021keeping} & ImageNet-21K & $16$ & $958.8 \times 10 \times 3$ & $81.1$ & $95.2$ \\
     MorphMLP-B~\cite{zhang2021morphmlp} & ImageNet-1K & $32$ & $197 \times 1 \times 4$ & $80.8$ & $94.9$ \\
    PST-B~\cite{xiang2022spatiotemporal} & ImageNet-21K & $32$ & $252 \times 4 \times 3$ & $82.5$ & $95.6$ \\
     Swin-B~\cite{liu2022video} & ImageNet-21K & $32$ & $282 \times 4 \times 3$ & $82.7$ & $95.5$ \\ 
  SIFA-Transformer~\cite{long2022stand} & ImageNet-21K & $32$ & $270 \times 4 \times 3$ & $83.1$ & $95.7$ \\
     DTF-Transformer~\cite{long2022dynamic} & ImageNet-21K & $32$ & $266 \times 4 \times 3$ & $\mathbf{83.5}$ & $95.9$ \\
     MViT-B~\cite{fan2021multiscale} & N/A & $64$ & $455 \times 3 \times 3$ & $81.2$ & $95.1$  \\
     MLP-3D-L~\cite{qiu2022mlp} & ImageNet-1K & $64$ & $308 \times 4 \times 3$ & $81.4$ & $95.2$ \\
     TimeSformer-L~\cite{bertasius2021space} & ImageNet-21K & $96$ & $7140 \times 1 \times 3$ & $80.7$ & $94.7$ \\
     Unifomer-B~\cite{liuniformer} & ImageNet-1K & $32$ & $259 \times 4 \times 3$ & $83.0$ & $95.4$ \\
    \hline 
     EMIM-S & ImageNet-1K & $16$ & $85 \times 4 \times 1$ &  $81.3$ & $94.9$ \\
    EMIM-B & ImageNet-1K & $16$ & $198 \times 4 \times 3$ & $82.7$ &  $95.5$\\
    EMIM-B & ImageNet-1K & $32$ & $526 \times 4 \times 3$ & $83.4$ & $\mathbf{96.1}$ \\
    \bottomrule[1pt] 
    \end{tabular}
    }
    \caption{Comparison with state-of-the-art methods on Kinetics-400}
    \label{Kinetics-400_result}
\end{table*}

\textbf{Evaluation on Kinetics-400.}
We also compare our proposed method with a set of state-of-the-art methods on Kinetics-400, which is a scene-related dataset. As shown in Table~\ref{Kinetics-400_result}, our method performs better than most of the state-of-the-art methods, except for DTF-Transformer, i.e., $83.4\%$ v.s. $83.5\%$. 
This is mainly because the performance of Kinetics-400 is highly related to the pretrained model, where DTF-Transformer is based on ImageNet-21K pretrained model, and our method is trained on ImageNet-1K pretrained model.
It is noticeable that, even with ImageNet-1K pretrained model, our method still achieves competitive results compared with other methods accompanying with ImageNet-21K model. This factor shows both the efficiency and effectiveness of our method in boosting performance.
In practice, we notice that the performance improvements on Kinetics-400 are not as significant as those on SSv1 \& SSv2. The main reason is that the scene-related dataset of Kinetics-400 has less motion information compared with motion-sensitive datasets. Under this circumstance, the maximum capacity of our method in motion modeling would be limited. Similar observations have also been discussed in other motion-dedicated methods~\cite{li2020tea,zhuang2022action}.
We leave it as an open problem in our work, and hope to explore it in the future.

\subsection{Ablation Studies}
\label{section_ablation_study}
In this section, we conduct extensive ablation experiments to verify the effectiveness of our proposed designs. Unless stated otherwise, we choose Uniformer-small as our baseline model and perform ablation investigations on  SSV1 with a 16-frame setting. For fair comparisons, we also remove the temporal downsampling strategy in the original Uniformer to preserve more temporal information.

\begin{table}[t]
    \centering
    \begin{tabular}{c|c|c|c|c|c}
    \hline \hline 
     Method   & Pretrained & Flops & Inference Time  & Top1 & Top5  \\ \hline
      Baseline  & ImageNet-1K & $110$G  & $10$ ms & $57.1$ & $85.0$ \\
      Our Method & ImageNet-1K  & $85$G & $14$ ms & $58.8$ & $85.7$ \\ 
      \hline \hline
      Baseline & Kinetics-400 & $110$G  & $10$ ms & $58.7$ & $85.6$ \\
      Our Method & Kinetics-400 & $85$G &  $14$ ms  & $59.9$ & $86.5$ \\
      \hline \hline
    \end{tabular}
    \caption{Performance comparison between our method and baseline method. The performance results with different pretrained models are shown.}
    \label{tab:baseline_ssv1}
\end{table}

\textbf{Comparison between Our Method and Baseline Method:} As shown in Table~\ref{tab:baseline_ssv1}, we compare our method with the baseline method based on different pretrained models, e.g. ImageNet-1K and Kinetics-400.
Note that the Kinetics-400 pre-trained models, either for the baseline method or our method, are trained by removing the temporal downsampling strategy for fair comparisons.
It is clear that our method performs well in comparison to the baseline method by a large margin, with $1.7\%$ and $1.2\%$ performance improvements on corresponding pre-trained models.
This factor indicates the effectiveness of our proposed designs in enriching motion modeling capabilities and boosting the performance of motion-sensitive datasets.
Furthermore, since the performance is consistently improved with different pre-trained models, it also shows the robustness and generalization capacity of our proposed method.
Last but not least, 
our method also shares the merit of lighter computational complexities, in comparison with the baseline method, i.e. $85$G v.s. $110$G.
This factor shows the possibility of achieving effective motion modeling and efficient computation at the same time.
However, the inference time of our method is slightly longer than that of baseline method, i.e. 14 ms v.s. 10 ms. The inference latency stems from the sliding-window operation, where additional temporary cache data is generated and slows down CPU performance. The influence of this factor has also been noted in existing studies~\cite{hassani2022neighborhood}.
Note that, in terms of the experiment pipeline, comparisons based on Kinetics-400 pretrained models are more trivial than those based on ImageNet-1K, as we need to pretrain models with the data of Kinetics-400 first. In this case, for simplicity, we perform the following comparisons based on ImageNet-1K.

\begin{table}[t]
    \centering
    \begin{tabular}{c|c|c|c}
    \hline \hline 
       Neighborhood Size  & Flops & Top1 & Top5 \\ \hline
       $3 \times 3$  & $84.7$G &  $57.5$ & $85.0$ \\
       $5 \times 5$ & $85.2$G & $57.8$ & $85.3$ \\
       $7 \times 7$  & $86.2$G & $58.6$ & $85.6$ \\
    $9 \times 9$  & $88.0$G & $58.4$ & $85.4$ \\
       $11 \times 11$  & $91.0$G & $58.4$ & $85.1$ \\
       \hline \hline
     \end{tabular}
    \caption{Investigation on the size of neighborhood}
    \label{tab:ablation_on_neighborhood_size}
\end{table}

\begin{table}[t]
    \centering
    \begin{tabular}{c|c|c}
    \hline \hline 
  Sampling Strategy & Top1 & Top5 \\ \hline
    Non-sliding Window  &  $58.3$ & $85.2$ \\
   Sliding Window   &  $58.8$ & $85.7$ \\ \hline \hline
    \end{tabular}
    \caption{Investigation on token sampling strategy.}
    \label{tab:sampling_strategy}
\end{table}

\textbf{Investigation on the Size of Neighborhood:} In order to investigate the influence of the neighborhood size,
we gradually change this factor, i.e. from $3 \times 3$ to $11 \times 11$.  
Note that, since the spatial resolution of Stage 3 and Stage 4 is $14 \times 14$ and $7 \times 7$, we only replace our EMIM module in Stage 3 for this ablation study. As shown in Table~\ref{tab:ablation_on_neighborhood_size},
with the growth of the neighborhood size, performance is consistently improved but reaches its maximum with the size of $7 \times 7$. Performances with a larger neighborhood size even slightly drop, i.e. from $58.6\%$ to $58.4\%$.
The main reason is that the receptive field of the $7 \times 7$  neighborhood covers most of the area of a frame and may be large enough to capture most movements. 
Large neighborhood sizes may bring no benefits for motion capturing and may even introduce irrelevant or harmful information for motion modeling. In this case, 
we set the default neighborhood size as $7 \times 7$.

\textbf{Investigation on Token Sampling Strategy:} 
We conduct different token sampling strategies to investigate the benefit of the explicit geometric inductive bias.
To identify the benefit of the explicit geometric inductive bias brought by the adopted token sampling strategy, we perform ablation studies with different token sampling strategies. 
For the baseline model, we adopt the non-sliding window sampling strategy, where key tokens are sampled in a query-irrelevant manner.
In contrast, our method adopts the sliding window sampling strategy, where the set of key candidate tokens is adaptively sampled from the query-based neighboring area.
As shown in Table~\ref{tab:sampling_strategy},
the model with the sliding window sampling strategy performs better than that with the non-sliding window sampling strategy,
i.e. $58.8\%$ v.s. $58.3\%$.
This is because the definite and well-aligned offsets and corresponding probabilities obtained by the sliding window sampling strategy can serve as a kind of explicit motion pattern, indicating how an object would move in the next frame. This factor would contribute a lot to effective motion modeling.

\begin{table}[]
    \centering
    \begin{tabular}{c|c|c}
    \hline \hline 
    Motion Feature & Top1 & Top5 \\ \hline
    without & $58.1$ & $85.3$\\
     with   & $58.8$  & $85.7$ \\ \hline \hline
    \end{tabular}
    \caption{Investigation on block replacement strategy}
    \label{tab:investigation_on_motion}
\end{table}

\textbf{Investigation on Motion Feature:}
To verify the benefit of displacement field information encoding, we attempt to remove the motion feature in our method. This means that the constructed affinity matrix will only be used for contextual aggregation and the displacement field information preserved in that affinity will be discarded. As shown in Table~\ref{tab:investigation_on_motion}, our method performs better than the baseline method, i.e. $58.8\%$ v.s. $58.1\%$. 
This is because contextual aggregation is a kind of weighted sum of candidate tokens for appearance modeling, where no displacement information of candidate tokens over anchor query tokens would be presented in the resulting appearance features. Alternatively, our method resorts to encoding the explicit offset information, i.e., visual similarities or probabilities, within the affinity matrix into motion features for motion modeling, which is complementary to the aforementioned appearance modeling.

\begin{table}[]
    \centering
    \begin{tabular}{c|c|c|c}
    \hline \hline
    Block Replacement Strategy & Flops & Top1 & Top5 \\ \hline
     E-E & $62$G & $55.8$ & $83.7$ \\
     O-O     & $110$G & $57.1$  & $85.0$ \\
      O-E    & $86$G & $58.2$ & $85.6$ \\ 
    E-O    & $85$G & $58.8$ & $85.7$ \\ 
     \hline \hline
    \end{tabular}
    \caption{Investigation on block replacement strategy. O denotes the \textbf{O}riginal self-attention module, while E denotes the proposed \textbf{E}MIM module.}
    \label{tab:block_replacement_strategy}
\end{table}

\textbf{Investigation on Block Replacement Strategy:} To explore the effect of different block replacement strategies, we provide four replacement strategies to examine this factor. For every transformer block in two successive transformer blocks, we alternatively choose whether to replace the \textbf{O}riginal self-attention module  with our \textbf{E}MIM module or keep it unchanged. As shown in Table~\ref{tab:block_replacement_strategy}, 
the E-O strategy performs best, where the self-attention module is replaced in the first block and the second block is unchanged.
Second, either the E-O strategy or the O-E strategy performs better than the baseline O-O strategy, while the E-E strategy performs the worst.
Such a replacement strategy could also be observed in SIFA~\cite{long2022stand} and DTF~\cite{long2022dynamic}.
As the pretrained ImageNet-1K model is trained with the original self-attention module, massive replacement of the original module on the downstream task can destroy the original weight distributions and lead to performance drop, that is, from $57.1\%$ to $55.8\%$. 
In this case, it is recommended to replace the original self-attention module with our EMIM to improve motion modeling capacity, but not too much, and to reduce optimization instability during the early stage of training.
We empirically set the replacement strategy as the E-O strategy.

\begin{table}[]
    \centering
    \begin{tabular}{c|c}
    \hline \hline
       Model  &  Top-1 Accuracy\\ \hline
       Uniformer-S  & $82.8$ \\ 
       EMIM-S & $82.9$  \\
       \hline \hline
    \end{tabular}
    \caption{Investigation on contextual aggregation capacity preserved.}
    \label{tab:contextual_aggregation_capacity}
\end{table}

 \begin{table}[t]
    \centering
    \begin{tabular}{c|c}
    \hline \hline
       Model  &  Top-1 Accuracy\\ \hline
       Traditional Method  & $57.9$ \\ 
       Our Method & $58.8$  \\
       \hline \hline
    \end{tabular}
    \caption{Comparison with traditional cost volume integration method.}
    \label{tab:comparison_with_traditonal_cost_volume}
\end{table}

\noindent\textbf{Investigation on Contextual Aggregation Capacity Preserved.}
To investigate the contextual aggregation capacity preserved in our design,
we remove the motion feature from our EMIM module, and degrade the application scenario of our EMIM module to static images, where key and value tokens are sampled from the same frame. 
We conduct the experiment on ImageNet-1K with our EMIM-S model.
 As shown in Table~\ref{tab:contextual_aggregation_capacity}, our EMIM-S achieves $82.9\%$ top-1 accuracy on ImageNet-1K, which is the same as or even slightly better than the performance of its counterpart Uniformer-S, i.e., $82.8\%$. 
 This factor validates that our EMIM module remains the original contextual aggregation capacities despite the proposed designs in EMIM.

 \noindent \textbf{Comparison with Traditional Cost Volume Integration Method.} We also try to integrate the traditional cost volume feature into the plain transformer-based method for motion modeling, i.e., Uniformer-S. In this case, we do not modify the original self-attention. The top-1 accuracy of the traditional method is smaller than that of our method, i.e. $57.9\%$ v.s. $58.8\%$. This factor indicates it is suboptimal to directly integrate the traditional cost volume feature into existing transformer-based methods in a naive way. Since we propose to ingeniously integrate those critical properties of cost volume into self-attention in a unified way, cost volume and self-attention could be collaboratively optimized and work together to boost performance.

\begin{table}[t]
    \centering
    \begin{tabular}{c|c|c} 
    \hline \hline 
    Temporal Interval & Top1 & Top5 \\ \hline
    1 (default setting)    & $58.8$ & $85.7$ \\
     2  & $58.0$ & $85.0$ \\
    3 & $56.9$  & $84.9$  \\ \hline \hline
    \end{tabular}
    \caption{Investigation on temporal interval.}
    \label{temporal_interval}
\end{table}

\begin{figure*}[t]
    \centering
    \includegraphics[width=1.0\textwidth, height=0.80\textwidth]{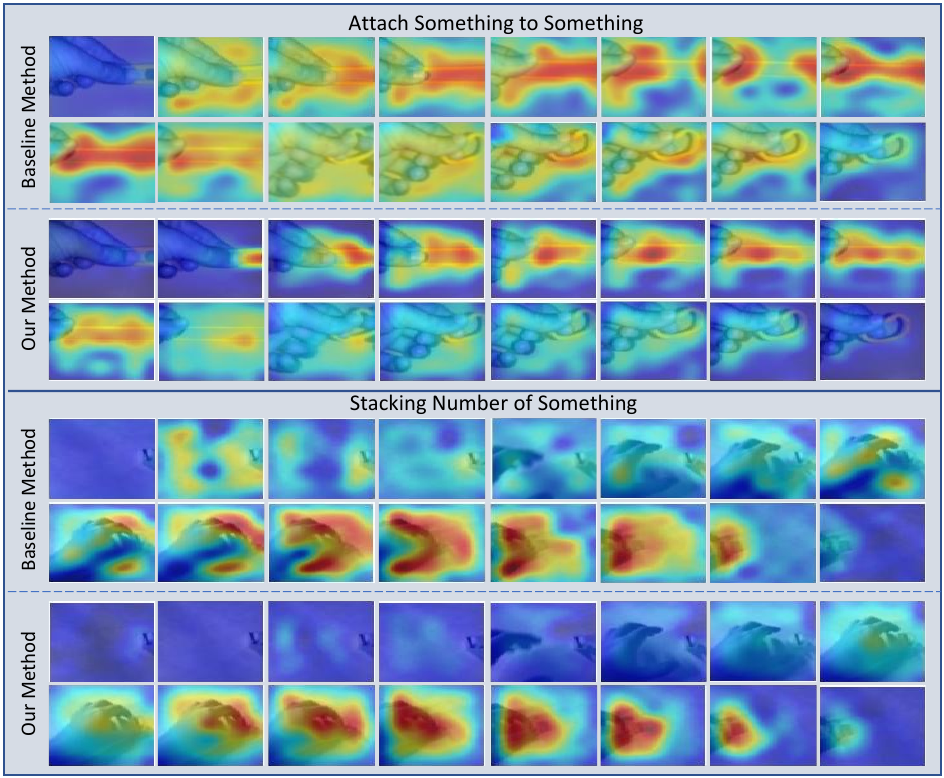}
    \caption{Visualization. To verify the effectiveness of our method in motion modeling, we select two action instances from SSv1 to show regions of interest of models via the Grad-CAM technique~\cite{selvaraju2017grad}. 
    Based on that, we have two important observations. First, for the first few still frames, there are no obvious movements between those frames. Under this circumstance, only a little attention is paid to the first few frames in our method, while a large portion of areas is mistakenly captured by the baseline method.
    Second, for the remaining quickly moving frames, our method pays finer and more accurate attention to the moving objects, compared with the baseline method.
    Based on these findings, it is clear that our method shares the powerful capacity of motion sensitivity and motion capturing. 
    (Best viewed when in color)
    }
    \label{fig:visualization}
    \vspace{-0.5cm}
\end{figure*}

\noindent \textbf{Investigation on Temporal Interval.} To investigate the influence of the temporal interval, we extend the temporal interval from $1$ to $3$ to investigate the influence of this factor. As shown in Table~\ref{temporal_interval}, our method achieves the best performance under the default setting, where the temporal interval is set to 1. As the interval increases, the performance gradually decreases. The inconsistent visual changes caused by large temporal intervals may introduce irrelevant information or even harmful information for motion modeling, and consequently deteriorate the performance. In this case, we set the temporal interval as 1 by default.

\section{Visualization}
\label{visualization}

To show the effectiveness of our method in motion modeling, we select two action instances from SSv1 for qualitative investigation. Those action instances are correctly recognized by our method, but incorrectly recognized by the baseline method. We show regions of interest of different models via the Grad-CAM~\cite{selvaraju2017grad} technique, where the important regions contributing to the final prediction would be highlighted.

As shown in Figure~\ref{fig:visualization}, we have two important observations.
First, for the first few still frames, there are no obvious movements between those frames. 
Under this circumstance,
it means that too much attention paid to these frames would introduce irrelevant information for accurate recognition, and even deteriorate the performance.
As a result,
 only a little attention is paid to the first few frames in our method, while a large portion of areas is mistakenly captured by the baseline method.
    Second, for the remaining quickly moving frames, our method pays finer and more accurate attention to the moving objects, compared with the baseline method.
    Based on these findings, it is clear that our method shares the powerful capacity of motion sensitivity and motion capturing, and consequently boosts the performance.

\section{Conclusion}
In this paper, we propose an Explicit Motion Information Mining module (EMIM) to enrich the motion modeling capability of existing transformers in action recognition.
Specifically, the EMIM module is achieved by integrating three critical properties of cost volume, e.g., local neighborhood, explicit geometric inductive bias, and displacement field information encoding, when the affinity matrix is constructed in self-attention.
We validate the effectiveness of our proposed designs on four widely-used datasets, and achieve state-of-the-art results on those datasets, especially on motion-sensitive datasets, i.e. SSV1 \& SSV2. In this work, we only exploit the native implementation of integrating the key properties of cost volume into the affinity matrix to improve motion capacity, without investigating its advanced potential. In the future, it could be further improved by incorporating the merits of traditional local feature descriptors in video understanding and integrating temporal trajectories to enhance long-term motion modeling capacities. Besides, it could also be integrated into other architectures for spatiotemporal tasks, such as the encoder in 3D-VAE and DiT~\cite{peebles2023scalable}.

\section{Limitations}
While we demonstrate that integrating three key cost volume properties into the affinity matrix could enhance the motion modeling capacity of existing transformers, there remain several valuable aspects that need to be explored. In our work, we reformulate the original affinity matrix as a displacement field, i.e., offsets and corresponding similarities/probabilities, and encode this information into motion representations. Given the conceptual similarity between the affinity matrix and traditional local feature descriptors in video understanding~\cite{dalal2005histograms,lowe2004distinctive}, it is promising to reutilize the informative displacement field through careful designs to better align with the characteristics of these classical descriptors. Second, we focus on using the motion information encoded in the affinity matrix for short-term motion modeling. However, it is also promising to combine the affinity matrix with temporal trajectories to construct a new trajectory representation~\cite{wang2015action,wang2013action} that enhances long-term motion modeling capabilities.

\bibliographystyle{elsarticle-num}
\bibliography{reference} 
\end{document}